\documentclass{article}

\usepackage[preprint]{neurips_2023}

\usepackage{natbib}
\setcitestyle{numbers,square}

\usepackage[utf8]{inputenc} 
\usepackage[T1]{fontenc}    
\usepackage{hyperref}       
\usepackage{url}            
\usepackage{booktabs}       
\usepackage{amsfonts}       
\usepackage{nicefrac}       
\usepackage{microtype}      
\usepackage{xcolor}         
\usepackage{multirow}
\usepackage{graphicx}
\usepackage{authblk}
\usepackage{grffile}
\usepackage{mwe}

\title{What Makes for Good Visual Tokenizers for \\Large Language Models?}

\author{
\textbf{Guangzhi Wang$^{1,2}$\thanks{This work is done during Guangzhi's internship at ARC Lab, Tencent PCG} \ \ Yixiao Ge$^{2,3}$\thanks{Project lead.} \ \ Xiaohan Ding$^{3}$\ \ Mohan Kankanhalli$^{1}$ \ Ying Shan$^{2,3}$} 

$^{1}$National University of Singapore\ \ $^{2}$ARC Lab, Tencent PCG\ \
$^{3}$Tencent AI Lab 
}

\begin{document}

\maketitle

\begin{abstract}
    We empirically investigate proper pre-training methods to build good visual tokenizers, making Large Language Models (LLMs) powerful Multimodal Large Language Models (MLLMs).
    In our benchmark, which is curated to evaluate MLLM's visual semantic understanding and fine-grained perception capabilities, we discussed different visual tokenizers pre-trained with dominant methods (\textit{i.e.}, DeiT, CLIP, MAE, DINO), and observe that:  i) Fully/weakly supervised models capture more semantics than self-supervised models, but the gap is narrowed by scaling up the pre-training dataset. 
    ii) Self-supervised models are better at fine-grained perception, where patch-level supervision is particularly effective.
    iii) Tuning the visual tokenizer leads to the loss of semantics obtained from large-scale pretraining, which is unfavorable with relatively small-scale instruction-tuning dataset.
    Given the findings, we reviewed methods that attempted to unify semantics and fine-grained visual understanding, \textit{e.g.}, patch-level feature distillation with semantically-rich targets. We obtain an intriguing insight: \textit{mask-based strategies that were once all the rage may not be applicable for obtaining good visual tokenizers}. 
    Based on this critical observation, we obtain a new MLLM equipped with a tailored Good Visual Tokenizer -- GVT, which exhibits strong visual comprehension capability at multiple scales.
    In particular, without introducing extra parameters and task-specific fine-tuning, GVT achieves superior performance on visual question answering, image captioning, and other fine-grained visual understanding tasks such as object counting and multi-class identification.
    Project released at: https://github.com/TencentARC/GVT
\end{abstract}

\section{Introduction}
Large Language Models (LLMs)~\cite{brown2020language_gpt3, touvron2023llama, radford2019language_gpt2, ouyang2022training_instructGPT} have demonstrated remarkable performance for various downstream tasks without task-specific fine-tuning.
Recently, based on the powerful LLMs, there has been a surge of research~\cite{li2023blip2, alayrac2022flamingo, zhu2023minigpt4, liu2023visual_llava, ye2023mplugowl, huang2023language_kosmos, yang2023mmreact, driess2023palme} that successfully adapt LLMs to vision-language tasks, resulting in powerful Multimodal LLMs (MLLMs), \textit{e.g.}, BLIP-2~\cite{li2023blip2}.
When properly fed with visual data, they are shown to be capable of understanding the visual world and responding to instructions accordingly.
Such vision-language understanding capability makes LLM a universal interface for multimodal tasks, contributing towards a tentative yet promising direction towards Artificial General Intelligence (AGI)~\cite{bubeck2023sparks, openai2023gpt4}.



\begin{figure}
    \centering \includegraphics[width=1.0\textwidth]{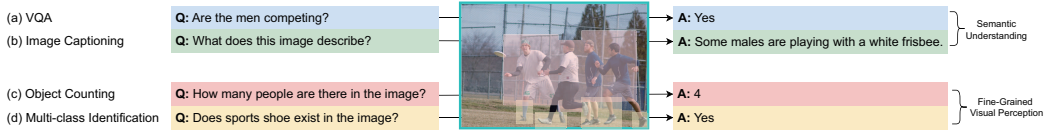}
    \caption{
    Different tasks require visual understanding of different perspectives.
    Mainstream vision-language tasks, \textit{e.g.}, (a) VQA and (b) Image Captioning mainly focus on semantic understanding of the image.
    In this work, we also study two fine-grained visual understanding  tasks: (c) Object Counting (OC) and (d) Multi-Class Identification (MCI).
    }\label{fig:teaser}
\end{figure}

Within this framework, images are projected to the linguistic space for the LLMs to understand, where the common practice employs an image-text pre-trained visual tokenizer\footnote{In this work, we study visual tokenizers which map images into a continuous latent space.}, \textit{i.e.}, CLIP~\cite{openclip}. 
However, even though CLIP has shown strong capacity for image representations, to the best of our knowledge, \textit{it is yet to be explored whether CLIP is the optimal visual tokenizer for MLLMs}.
The absence of such investigation calls for a comprehensive comparison of existing visual tokenizers under MLLMs' framework. 
However, 
recent MLLMs have mostly investigated their performance in terms of generation quality~\cite{zhu2023minigpt4, liu2023visual_llava} or on a small set of questions~\cite{ye2023mplugowl}, leaving a comprehensive quantitative evaluation untouched.


To this end, we curated a new benchmark to study what makes for a Good Visual Tokenizer (GVTBench). 
It is especially designed to evaluate an MLLM's visual understanding capability from two important perspectives: semantic understanding and fine-grained visual perception capabilities.
As shown in Figure~\ref{fig:teaser}, the former is evaluated on Visual Question Answering (VQA) and image captioning.
While the latter is tested on two new tasks: Object Counting (OC) and Multi-Class Identification (MCI), which requires in-depth understanding of fine-grained visual information.
Based on this benchmark, we comprehensively evaluated existing visual tokenizers with same architecture but different pretraining methods, including fully supervised (DeiT~\cite{touvron2021training_deit}), text-guided weakly supervised (CLIP~\cite{radford2021learning_clip}) and self-supervised (MAE~\cite{he2022masked_mae}, DINO~\cite{ caron2021emerging_dino}, DINOv2~\cite{oquab2023dinov2}) models (Section~\ref{sec:empericial_study}). 
Our main observations are 
\textbf{i)} fully supervised and text-guided weakly supervised visual tokenizers demonstrate better semantic representation capacity than their self-supervised counterparts, but the gap is narrowed by scaling up the pretraining dataset (\textit{i.e.}, CLIP \textit{vs.} DINOv2). 
\textbf{ii)} Self-supervised visual tokenizers show better fine-grained visual perception capacity, where patch-level supervision leads to superior region-level understanding. 
\textbf{iii)} On instruction tuning datasets which are often smaller than visual tokenizer pretraining dataset~\cite{liu2023visual_llava, zhu2023minigpt4}, jointly tuning the visual tokenizer leads to noticeable semantic loss (\textit{i.e.}, frozen CLIP performs much better than tunable CLIP on semantic understanding tasks).


Given the fact that none of the previous visual tokenizers exhibit both good semantic and fine-grained visual perceptual capabilities, we reviewed existing methods that integrate semantic and region supervision and question whether they bring the best of the two worlds into a visual tokenizer. 
Existing methods can be mainly divided into two categories.
Methods in the first group~\cite{zhong2022regionclip, minderer2022simple_owlvit} enhance a pretrained CLIP with region-level supervision, which comes from a pretrained Region Proposal Network (RPN) or bounding box annotations.
However, we found that this leads to the loss of original semantics, which can not be justified by the limited improvements on fine-grained visual perception capabilties.
The other group of methods~\cite{eva, wei2022FD} utilize patch features from a pretrained CLIP as region supervision to train a new model, intending to enhance its fine-grained visual perceptual capability while maintaining the rich semantics.  
Specifically, ~\cite{eva, wei2022mvp} uses CLIP features to supervise the training of Masked-Image-Modeling (MIM), while Feature Distillation~\cite{wei2022FD} directly distills the CLIP feature into a new model without patch masking.
Nonetheless, the introduction of \texttt{[MASK]} token in MIM leads to train-test mismatch, requiring the visual tokenizer to be jointly optimized in the instruction-tuning process, which again leads to semantic loss with the small-scale instruction tuning dataset.
As such, we argue that the mask-based strategies that were once all
the rage may not be applicable for obtaining good visual tokenizers under MLLM's framework.

Based on these insights, we seek a new visual tokenizer with both strong semantic understanding and fine-grained visual perception capabilities via Feature Distillation~\cite{wei2022FD}.
Specifically, given a pretrained CLIP with rich semantics, we distill it into a new model by using the patch features as supervision, without patch masking.
In this way, the rich semantics from large-scale image-text contrastive pretraining is preserved, and the fine-grained visual perceptual capability is greatly enhanced with patch supervision.
With our new visual tokenizer and the language model Vicuna~\cite{vicuna}, we obtain a new MLLM with \textbf{G}ood \textbf{V}isual \textbf{T}okenizer (GVT).
Benefiting from the versatile visual tokenizer, GVT is able to perform well vision language tasks that require visual understanding at multiple levels.
Without introducing extra parameters, we achieve superior performance on semantic understanding tasks, \textit{i.e.}, VQA and image captioning, as well as fine-grained visual understanding tasks: instance counting and multi-class identification.

To summarize, our contributions are as follows:
\begin{itemize}
    \item To effectively evaluate MLLM's visual understanding capacity at different levels, we curate a new benchmark (GVTBench) which includes both semantic understanding tasks (VQA and image captioning) as well as fine-grained visual understanding tasks (Object Counting and Multi-Class Identification). 
    Based on GVTBench, we perform extensive experiments to study what makes for a good visual tokenizer for MLLMs and make three main observations.

    
    \item We reviewed methods that combine CLIP with fine-grained supervision to see if they can achieve the best of both worlds in terms of visual semantics and fine-grained understanding. We found that the SOTA pre-trained models (\textit{i.e.}, EVA) are inapplicable due to the train-test mismatch caused by MIM. Such mask-based visual tokenizers rely on further tuning with instructions,
    which leads to the loss of pre-trained rich semantics.

    
    \item 
    Based on the insights, we tailor a new visual tokenizer by distilling the patch-level semantics of a pre-trained CLIP without masking.
    With our visual tokenizer and Vicuna~\cite{vicuna}, we arrive at a superior MLLM (GVT) with strong visual understanding capability, achieving state-of-the-art performance on our curated benchmark.
\end{itemize}

\section{GVTBench for Empirical Study}\label{sec:empericial_study}
To comprehensively study what makes for good visual tokenizers for MLLMs, we conduct a series of experiments to study the property of various visual tokenizers with same architecture but different pretraining methods.
In this work, we mainly investigate MLLMs' visual understanding capability from two important perspectives: semantic understanding and fine-grained visual perception.

\subsection{Experimental Setup}

\noindent\textbf{GVTBench.}
A comprehensive evaluation requires a benchmark that suitably quantify MLLM's visual understanding capability.
Nonetheless, existing vision-language tasks mainly focus on semantic understanding~\cite{farhadi2010every, goyal2017making_vqav2}, leaving a special focus on fine-grained visual perception untouched.
To this end, we curated a new benchmark -- GVTBench. 
It evaluates the semantic understanding capability of an MLLM on VQA~\cite{goyal2017making_vqav2} and Image Captioning (IC)~\cite{lin2014microsoft_coco}. 
We report accuracy for the former and CIDEr~\cite{cider} and SPICE~\cite{spice} for the latter. 
For fine-grained visual perception capability evaluation, we specially design two new tasks for MLLMs:
\begin{itemize}
    \item \textbf{Object Counting (OC)}. We ask the model to count the number of a certain object appearing in the image with the prompt \textit{``Question: How many \{obj\}  are there in the image? Answer:''}. 
    We regard it as a classification task and report a model's prediction accuracy.
    \item \textbf{Multi-Class Identification (MCI).} We ask the model if a certain object exists in the image with the prompt \textit{``Question: Does \{obj\} exist in the image? Answer:''}. The model is expected to answer \textit{`` Yes/No''}, resulting in a binary classification problem. We report accuracy for this task. 
\end{itemize}

Notably, in the VQAv2~\cite{goyal2017making_vqav2} benchmark, there are also questions related to numbers. 
Nevertheless, these questions are often coupled with high-level semantics, making it unsuitable to strictly evaluate fine-grained visual understanding capabilities.
In contrast, our OC and MCI tasks attend to individual objects, which is decoupled from high-level semantics and thus a more appropriate test bed for fine-grained visual understanding evaluation.

\noindent\textbf{Experimental Setting.}
We use different visual tokenizers to encode an image into a set of visual tokens.
Then, we follow Flamingo~\cite{openflamingo} to use the Perceiver Resampler~\cite{jaegle2021perceiver} to reduce the number of visual tokens to a fixed length, which are fed into LLM (\textit{i.e.}, Vicuna).
The models are trained on a instruction dataset which contains about 5M image-text pairs.
In the training process, the language model is always frozen, while the visual tokenizer can be frozen or jointly optimized. 
For more implementation details, please refer to the appendix.

\begin{table}[htbp]
    \centering
    \caption{Comparison of visual tokenizers of ViT-B with different pretraining strategies. The \textbf{best} result is \textbf{bold} while \underline{the second best} is \underline{underlined}.}\label{tab:study}
    {\small
    \begin{tabular}{cclccccccc}
         \toprule
         Joint & \multirow{2}{*}{Supervised}  & Visual  & \# Pretraining  & VQA & \multicolumn{2}{c}{Captioning} & OC & MCI & \multirow{2}{*}{Avg} \\
         Tuning & & Tokenizer & Images & Acc & CIDEr & SPICE & Acc & Acc &  \\
         \midrule
         \multirow{5}{*}{$\times$} & Fully & DeiT~\cite{touvron2021training_deit} & 1.28 M & 48.3 & 65.8 & 15.9 & 37.5 & 83.6 & 58.8\\
         \cmidrule(lr){2-10}
          & \multirow{3}{*}{Self} & DINO~\cite{caron2021emerging_dino} & 1.28 M & 50.1 & 45.0 & 13.5 & 46.5 & 80.8 & 55.6 \\
          & & MAE~\cite{he2022masked_mae} & 1.28 M & 48.4 & 37.3 & 11.8 & \textbf{47.5} & 82.7 & 53.4 \\
          & & DINOv2~\cite{oquab2023dinov2} & 142  M & \underline{51.3} & \underline{67.9} & \underline{16.1} & \underline{47.0} & 86.0 & \textbf{63.1} \\         
         \cmidrule(lr){2-10}
          & Weakly & CLIP~\cite{radford2021learning_clip} & 400  M & \textbf{52.2} & \textbf{69.3} & \textbf{16.6} & 42.5 & 86.0 & \underline{62.5} \\
          
         \midrule
         \multirow{5}{*}{\checkmark} & Fully & DeiT~\cite{touvron2021training_deit} & 1.28 M & 50.7 & 38.4 & 10.0 & 41.0 & 86.9 & 54.3 \\
         \cmidrule(lr){2-10}
          & \multirow{3}{*}{Self} & DINO~\cite{caron2021emerging_dino} & 1.28 M & 47.3 & 54.1 & 14.5 & 44.5 & 86.6 & 58.1 \\
          & & MAE~\cite{he2022masked_mae} & 1.28 M & 48.9 & 48.0 & 14.2 & \textbf{47.5} & \textbf{88.7} & 58.2\\
          & & DINOv2~\cite{oquab2023dinov2} & 142  M & 50.5 & 49.6 & 13.0 & 43.5 & 84.1 & 56.9\\         
         \cmidrule(lr){2-10}
          & Weakly & CLIP~\cite{radford2021learning_clip} & 400  M & 47.7 & 64.2 & 15.4 & 45.5 & \underline{88.0} & 61.4\\

          \bottomrule
    \end{tabular}
    }
\end{table}

\subsection{Comparing Visual Tokenizers}
On GVTBench, we evaluate visual tokenizers with the same architecture (ViT-B~\cite{dosovitskiyimage_vit}) but different pretraining strategies, including fully-supervised (DeiT~\cite{touvron2021training_deit}), self-supervised (DINO~\cite{caron2021emerging_dino}, DINOv2~\cite{oquab2023dinov2}, MAE~\cite{he2022masked_mae}) and text-guided weakly supervised (CLIP~\cite{radford2021learning_clip}) pretraining. 
Based on the results in Table~\ref{tab:study}, we arrive at the following conclusions.

\textbf{Fully/weakly supervised models capture
more semantics than self-supervised ones, but the gap is narrowed by scaling
up the pre-training dataset.}
With tokenizers pretrained on relative small-scale dataset (\textit{i.e.}, ImageNet-1k~\cite{russakovsky2015imagenet} with 1.28M images), DeiT demonstrates better image captioning performance (65.8 CIDEr) than self-supervised models DINO (45.0) and MAE (37.3), without jointly tuning the visual tokenizer.
However, with 142M images for pretraining, the self-supervised model -- DINOv2 outperforms the supervised DeiT on image captioning (67.9) and VQA (51.3), and is only inferior to CLIP which is pretrained with weak supervision from a large-scale dataset with 400M image-text pairs.
This indicates that supervision is beneficial for semantic representation capability, but this can also emerge from large-scale pretraining with self-supervision.



\textbf{Self-supervised models are better at fine-grained
perception, where patch-level supervision is particularly effective.}
On fine-grained visual understanding tasks, \textit{i.e.}, OC and MCI,  self-supervised models demonstrate consistently better performance than those with supervision. 
When they are jointly tuned on the instruction dataset, their OC and MCI performance are mostly boosted, indicating their fine-grained visual perception capability gets improved.
Among all the self-supervised models, MAE achieves the best performance, indicating the patch-based supervision is particularly effective for improving fine-grained visual understanding. 

\noindent\textbf{Tuning semantic-rich visual tokenizer leads to semantic loss on small-scale instruction tuning dataset.}
When the tokenizer is jointly optimized on the instruction tuning dataset, the rich semantics obtained from large-scale pretraining in CLIP and DINOv2 have noticeably dropped (\textit{e.g.}, CLIP VQA 52.2 $\rightarrow 47.7$ and DINOv2 captioning 67.9 $\rightarrow$ 49.6).
We conjecture this is due to the relatively small scale of our instruction dataset ($\sim$5M $\ll$ 142M). 
As such, for modern MLLMs that are often tuned on small-scale and high-quality instruction datasets~\cite{zhu2023minigpt4, liu2023visual_llava}, jointly tuning the visual tokenizer may not be a good option.
\section{Unifying Semantic and Fine-grained Visual Understanding}

\subsection{CLIP with Region-based Training}
The generalist MLLMs call for a versatile visual tokenizer that could properly represent an image's content at multiple levels. 
However, based on the results in Table~\ref{tab:study},  none of existing pretraining methods leads to a good visual tokenizer that excels at both semantic and fine-grained visual perception capabilities.
This motivates us to explore whether the best of the two worlds can be achieved by any other method. 

\begin{table}[h]
    \centering
    \caption{Comparison of Visual region supervised methods and CLIP.}\label{tab:region}
    {\small
    \begin{tabular}{clccccccc}
         \toprule
         Joint &  Visual  &   VQA & \multicolumn{2}{c}{Captioning} & OC & MCI & \multirow{2}{*}{Avg} \\
         Tuning & Tokenizer  & Acc & CIDEr & SPICE & Acc & Acc &  \\
         \midrule
         $\times$ & CLIP~\cite{radford2021learning_clip} & \textbf{52.2} & \textbf{69.3} & \textbf{16.6} & 42.5 & 86.0 & \textbf{62.5} \\
         $\times$ & RegionCLIP~\cite{zhong2022regionclip} & 48.7 & 28.5 & 10.3 & 41.0 & 86.0 & 51.5 \\
         $\times$ & Owl-ViT~\cite{minderer2022simple_owlvit} & 44.0 & 32.5 & 8.5 & 43.0 & 80.8 & 50.1 \\
         \midrule
         $\checkmark$ & CLIP~\cite{radford2021learning_clip} & 47.7 & 64.2 & 15.4 & 45.5 & \textbf{88.0} & 61.4 \\
         $\checkmark$ & RegionCLIP~\cite{zhong2022regionclip} & 49.7 & 65.5 & 14.1 & \textbf{47.5} & 86.4 & 62.3 \\
         $\checkmark$ & Owl-ViT~\cite{minderer2022simple_owlvit} & 50.8 & 61.2 & 14.0 & 38.5 & 87.1 & 59.4 \\
         \bottomrule
    \end{tabular}
    }
\end{table}

\noindent\textbf{Fine-tuning CLIP with region supervision.}
One stream of work~\cite{zhong2022regionclip, minderer2022simple_owlvit} attempted to improve region representation capability of a pretrained CLIP by fine-tuning it with region supervision, which has demonstrated improved performance for open-vocabulary object detection.
This motivates us to study if this also enhances CLIP as a visual tokenizer. 
We mainly investigated RegionCLIP~\cite{zhong2022regionclip} and Owl-ViT~\cite{minderer2022simple_owlvit}.
The former finetunes a CLIP with region-level supervision from bounding boxes generated by a pretrained RPN, while the latter utilizes the region annotation from an object detection dataset. 
We compared these methods with CLIP, and show the results in Table~\ref{tab:region}.
It can be observed that, without joint tuning the visual tokenizer, both RegionCLIP and Owl-ViT show severe performance drop on image captioning and VQA, indicating the rich semantics in the original CLIP is lost during their region fine-tuning process.
On the other hand, when the visual tokenizers are jointly tuned on the instruction-tuning dataset, their fine-grained representation capability improves by a margin (on OC and MCI performance), but this can not justify the loss of semantic representation capability, resulting in inferior overall performance compared to the original CLIP. 

\begin{table}[h]
    \centering
    \caption{Comparison of different strategies of utilizing CLIP features with ViT-B architecture.}\label{tab:mask}
    {\small
    \begin{tabular}{lclccccccc}
         \toprule
         \multirow{2}{*}{Method} & Joint &  Patch  &   VQA & \multicolumn{2}{c}{Captioning} & OC & MCI & \multirow{2}{*}{Avg} \\
         & Tuning & Masking  & Acc & CIDEr & SPICE & Acc & Acc &  \\
         \midrule
         CLIP~\cite{radford2021learning_clip} & $\times$ & - & \textbf{52.2} & 69.3 & \textbf{16.6} & 42.5 & 86.0 & 62.5 \\
         FD~\cite{wei2022FD}  &$\times$ & $\times$ & 49.4 & \textbf{72.1} & 15.8 & 46.5 & 86.7 & \textbf{63.7} \\
         EVA~\cite{eva} & $\times$ & $\checkmark$ & 42.9 & 27.0 & 10.0 & \textbf{46.9} & 70.5 & 46.8 \\
         \midrule
         CLIP~\cite{radford2021learning_clip} &$\checkmark$ & - & 47.7 & 64.2 & 15.4 & 45.5 & \textbf{88.0} & 61.4 \\
         FD~\cite{wei2022FD} & $\checkmark$ & $\times$ & 49.3 & 53.3 & 12.7 &  40.5 & 85.8 & 57.2 \\
         EVA~\cite{eva} & $\checkmark$ & $\checkmark$ & 51.4 & 61.6 & 12.3 & 45.9 & 87.1 & 61.5 \\
         \bottomrule
    \end{tabular}
    }
\end{table}

\begin{figure}[h]
    \centering
    \includegraphics[width=1.0\textwidth]{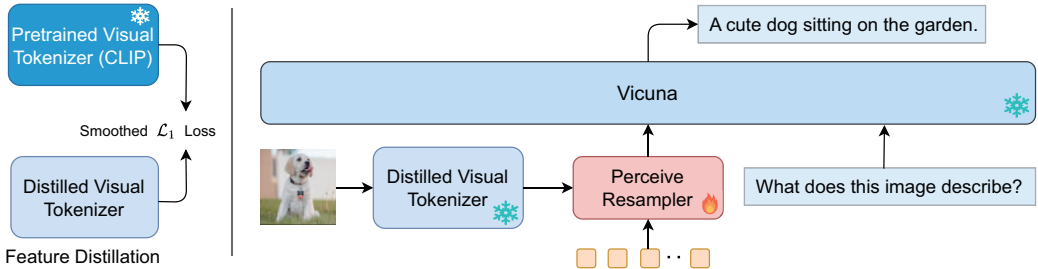}
    \caption{
    Framework of our GVT. 
    We first distill the features of a pretrained CLIP via smoothed $\mathcal{L}_1$ loss. 
    Then, we use it to encode images into a set of tokens, which are fed into the Perceiver Resampler~\cite{jaegle2021perceiver} as soft prompts. Together with language instructions, these prompts are fed into LLM to generate responses. Only the Perceiver Resampler is optimized in this process. 
    }\label{fig:gvt}
\end{figure}

\noindent\textbf{Semantic Feature as Region Supervision.}
Another stream of work utilized CLIP's patch feature as region-level supervision for pretraining, aiming to obtain a model with both strong semantics and better region representations.
Specifically, EVA~\cite{eva} and MVP~\cite{eva} use CLIP's patch feature as regression target for Masked Image Modeling (MIM) pretraining, while FD~\cite{wei2022FD} does not employ the masking strategy and directly distills CLIP's patch feature into a new model. 
We compared these methods in Table~\ref{tab:mask}.
Without jointly tuning the visual tokenizer, FD results in performance improvement on both semantic and fine-grained visual understanding upon CLIP.
However, when patch masking strategy is adopted, the performance of EVA significantly drops.
This can be attributed to the introduction of the \texttt{[MASK]} token for MIM, which is only used for pretraining the visual tokenizer but discarded afterwards.
In this way, the train-test mismatch arises without tuning the visual tokenizer, leading to unsatisfactory performance for downstream tasks. 
On the other hand, when the visual tokenizer is jointly optimized with the instruction data, they are inferior to the original CLIP on VQA and image captioning, indicating semantic loss occurs.


Given the fact that modern MLLMs are often trained on high-quality and small-scale instruction datasets~\cite{zhu2023minigpt4, liu2023visual_llava}, our observation suggests that visual tokenizer should be frozen to maintain the powerful semantic representation capability from large-scale pretraining.
Nonetheless, for visual tokenizers pretrained with MIM, the introduction of the \texttt{[MASK]} token inevitably leads to train-test mismatch, necessitating it to be jointly tuned on the instruction data.
This contradiction indicates that mask-based pretraining may not lead to a good visual tokenizer under MLLM's framework. 



\subsection{MLLM with Good Visual Tokenizer}
In this work, we tune a new visual tokenizer which unifies the advantage of semantic representation, fine-grained visual perception and semantic maintenance capabilities.
Based on the insights above, 
we achieve this objective by utilizing a visual tokenizer pretrained on large-scale datasets, and properly integrate it with patch-level supervision. 
Furthermore, we does not use any mask-based strategy, so as the rich semantics could be preserved by freezing it in the instruction tuning process.
Specifically, we take the powerful EVA-CLIP~\cite{EVA-CLIP} based on ViT-L as the teacher model, and randomly initialize another model with identical architecture as student.
The patch features from the teacher model is normalized by a whitening operation, and is taken as regression target for the student model.
Afterwards, the visual tokenizer can be used for MLLMs and kept frozen during instruction tuning.


%

Based on the tuned visual tokenizer, we construct a new MLLM with \textbf{G}ood \textbf{V}isual \textbf{T}okenizer (GVT).
The framework of GVT is shown in Figure~\ref{fig:gvt}.
Following~\cite{openflamingo}, we also random initialize a Receiver Resampler~\cite{jaegle2021perceiver} with 32 learnable queries to attend to the features from the visual tokenizer. 
Then, the features output from Perceiver Resampler are taken as soft prompts, and are fed into the LLM together with the language prompts.
In this work, we choose the instruction-tuned Vicuna-7B~\cite{vicuna} as the LLM.
The whole model is trained by the language modeling loss, and only the Perceiver Resampler is optimized in this process. 

\section{Experiments}

\subsection{Experimental Setup}
We train our model on a joint dataset of image-text pairs, including CC3M~\cite{sharma2018conceptual_gcc}, SBU~\cite{vicente2016large_sbu}, Visual Genome~\cite{krishna2017vg} and MS-COCO~\cite{lin2014microsoft_coco}.
We formulate these datasets as image captioning task, and use \textit{"what does the image describe?"} as prompt during training.
Besides, we also use two object detection datasets -- Object365~\cite{shao2019objects365} and OpenImagesV6~\cite{kuznetsova2020openimages} to design a set of object-centric tasks following~\cite{piergiovanni2022pre_objaware}.
The LLaVA-150k~\cite{liu2023visual_llava} dataset is also utilized for joint training.
This results in a total of 15M image-text pairs.
The images are resized to 224 $\times$ 224, and we adopt random resized crop and horizontal flipping for data augmentation during training.
The model is trained for 50k steps with 2k steps for linear warmup. 
We use AdamW~\cite{loshchilov2017decoupled_adamw} optimizer with a learning rate of 1e-4 and batch size 1024.
The training process takes about 2 days on 32 Tesla V100 GPUs.
For more implementation details, please refer to our appendix.

\subsection{Comparison with State-of-the-art Methods}
Without task-specific fine-tuning, we evaluate GVT on our GVTBench, which includes VQA~\cite{goyal2017making_vqav2}, Image Captioning~\cite{lin2014microsoft_coco}, Object Counting (OC) and Multi-Class Identification (MCI).
Besides evaluating OC and MCI on MS-COCO validation set, we also evaluate these two tasks based on the validation set of the VCR dataset~\cite{zellers2019recognition_vcr}.
We compared our method with recent MLLMs, including Flamingo~\cite{openflamingo}, BLIP-2~\cite{li2023blip2},
KosMos-1~\cite{huang2023language_kosmos}, LLaVa~\cite{liu2023visual_llava}, miniGPT4~\cite{zhu2023minigpt4}. 
We evaluate open-sourced models under our GVTBench and use reported results for others.
The results are shown in Table~\ref{tab:sota}.

On these tasks, our GVT achieves the best overall performance across competitors.
Specifically, on tasks requiring fine-grained visual perception, \textit{i.e.}, OC and MCI on both COCO and VCR datasets, GVT surpasses models with larger visual tokenizer and more curated data.
This indicates our visual tokenizer can better capture the fine-grained visual information, providing representations with better details.
For semantic understanding tasks, GVT achieves the second best with an accuracy of 60.4 on VQA.
This result is only inferior to BLIP-2, which utilized a much larger instruction dataset with high-quality image captions filtered by~\cite{li2022blip}.
On image captioning task, our GVT achieves the highest SPICE score and second best CIDEr, showing it also has strong semantic understanding capability.

\begin{table}[ht]
    \centering
    \caption{Comparison with State-of-the-arts. The \textbf{best} results are bold and the second best are \underline{underlined}.}\label{tab:sota}
    {\small
    \scalebox{0.85}{
    \begin{tabular}{lccccccccc}
         \toprule
         \multirow{2}{*}{Model}  & \#Vis. Tok.  & VQA & \multicolumn{2}{c}{COCO-Caption} & COCO-OC & COCO-MCI & VCR-OC & VCR-MCI & \multirow{2}{*}{Avg} \\
          & Params & Acc & CIDEr & SPICE                       & Acc     & Acc      & Acc    & Acc    \\
         \midrule
         Flamingo-9B~\cite{alayrac2022flamingo} & 438 M & 51.8 & 79.4 & - & - & - & - & - & -\\
         Kosmos-1~\cite{huang2023language_kosmos}&  307 M & 51.0 & 84.7 & 16.8 & - & - & - & - & -\\
         \midrule 
         LLaVa~\cite{liu2023visual_llava} & 307 M & 39.0 & 48.3 & 15.0 & 22.2 & 52.0 & 24.6 & 66.9 &  44.7 \\
         miniGPT4~\cite{zhu2023minigpt4} & 1.0 B & 58.2 & 80.6 & \underline{19.5} & 21.5 & 76.8 & \underline{25.1} & \underline{70.1} & 55.4 \\
         BLIP-2~\cite{li2023blip2} & 1.0 B & \textbf{62.4} & \textbf{93.3} & 17.3 & \underline{48.0} & \underline{81.9} & 20.2 & 68.9 & \underline{62.5}   \\
         \midrule
         GVT (Ours) & 307 M & \underline{60.4} & \underline{89.9} & \textbf{19.6} & \textbf{56.2} & \textbf{89.3} & \textbf{40.3} & \textbf{78.9} & \textbf{69.2} \\
         \bottomrule
    \end{tabular}
    }
    }
\end{table}

\subsection{Ablation Study}
We adopt the training protocol in Section~\ref{sec:empericial_study} to study the design of our GVT. 

\noindent\textbf{Choice of Distillation Target.}
According to the results in Table~\ref{tab:study}, we observe that DINOv2, which is pretrained with self-supervision on a dataset with 142M images also demonstrates good overall performance. 
To find the best target for feature distillation, we compared it with the CLIP model from~\cite{EVA-CLIP}, both in ViT-L architecture.
The results are shown in Table~\ref{tab:ab_tokenizer}.
It can be seen that CLIP has demonstrated better overall performance, which can be attributed to their large-scale pretraining dataset and advanced training strategies~\cite{EVA-CLIP}. 

\noindent\textbf{Number of Latent Queries.}
We study the number of latent queries in the Perceiver Resampler. 
The results are shown in Table~\ref{tab:ab_nquery}. 
It can be observed that the overall performance generally increases with the number of latent queries, where 32 query results in a satisfactory performance.
Besides, increasing the number of query to 64 leads to limited improvements.

\begin{table}[ht]
    \centering
         \caption{Comparison of visual tokenizers under ViT-L architecture.}
         \begin{tabular}{lccccccc}
         \toprule
         Visual &   VQA & \multicolumn{2}{c}{Captioning} & COCO-OC & COCO-MCI & \multirow{2}{*}{Avg} \\
         Tokenizer & Acc & CIDEr & SPICE & Acc & Acc &  \\
         \midrule
         DINO-v2-Large~\cite{oquab2023dinov2} & 53.9 & 69.9 & 15.0 & \textbf{45.5} & \textbf{83.6} & 63.2 \\
         CLIP-Large~\cite{EVA-CLIP}  & \textbf{55.5} & \textbf{71.9} & \textbf{16.5} & 45.2 & 83.5 & \textbf{64.0} \\
         \bottomrule
         \end{tabular}
    \label{tab:ab_tokenizer}
\end{table}

\begin{table}[ht]
    \centering
         \caption{Comparison of the number of latent queries in the Perceiver Resampler.}
         \begin{tabular}{cccccccc}
         \toprule
         \#Latent &   VQA & \multicolumn{2}{c}{Captioning} & COCO-OC & COCO-MCI & \multirow{2}{*}{Avg} \\
         Query & Acc & CIDEr & SPICE & Acc & Acc &  \\
         \midrule
         8  & 53.4 & 60.0 & 15.4 & 50.0 & 78.0 &  60.3 \\
         16 & 55.0 & 61.7 & 15.8 & \textbf{51.1} & 83.5 &  62.8 \\
         32 & \textbf{55.5} & \textbf{71.9} & \textbf{16.5} & 45.2 & 83.5 & 64.0 \\
         64 & 54.0 & 71.1 & 16.4 & 47.0 & \textbf{84.2} &  \textbf{64.1} \\
         \bottomrule
         \end{tabular}
    \label{tab:ab_nquery}
\end{table}

\subsection{Qualitative Results}
\begin{figure}[ht]
    \centering
    \includegraphics[width=1.0\textwidth]{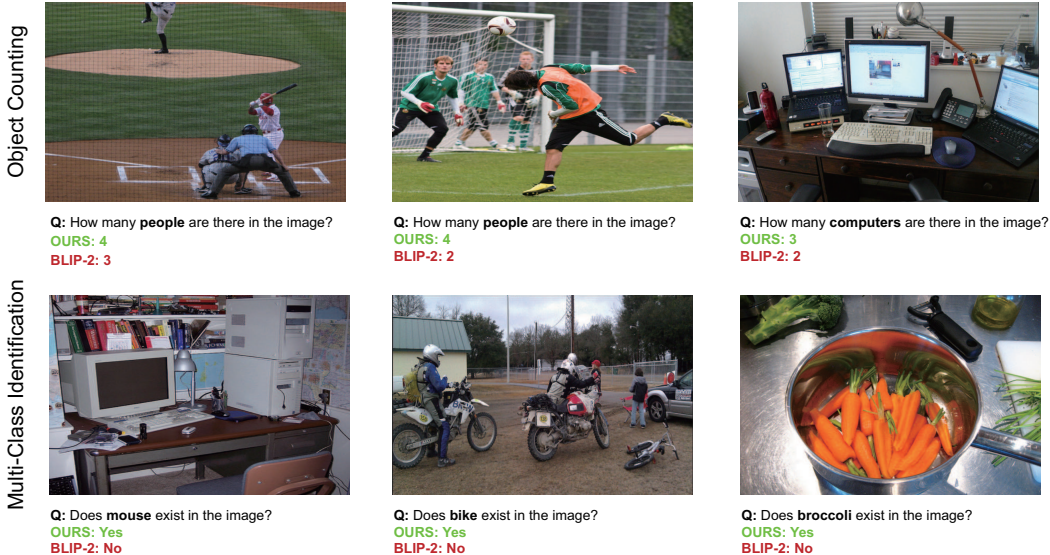}
    \caption{Qualitative Comparison on OC and MCI. Our method shows better performance on recognizing detailed clues in the image.}
    \label{fig:qualit_results}
\end{figure}

We show some qualitative comparison on OC and MCI between our GVT and BLIP-2 in Figure~\ref{fig:qualit_results}.
It can be observed that our method demonstrate better fine-grained visual understanding capabilities than the baseline method.
Take the first example in OC as an example, our method not only recognize the 3 people in the foreground, but also takes the fourth person who is far away from the camera into consideration.
Besides, GVT also successfully recognize non-salient or small-sized objects in the image, such as the mouse, bicycle and broccoli in the three examples in MCI.
\section{Related Work}

\subsection{Multimodal Large Language Models}
LLMs have demonstrated strong capabilities for various downstream tasks without task-specific fine-tuning. 
Based on this, recent work has utilized it to accomplish vision and language tasks, enabling powerful Multimodal Large Language Models.
The common practice uses a visual tokenizer to encode the image, followed by potential bridges such as MLP~\cite{liu2023visual_llava} or Perceiver Resampler~\cite{jaegle2021perceiver} to encode them into soft prompts.
For example, Flamingo~\cite{alayrac2022flamingo} adopts a contrastive pre-trained visual tokenizer, followed by a Perceiver Resampler~\cite{jaegle2021perceiver} to aggregate the image tokens into fixed length.
These tokens are fed into a frozen language model with the help of gated attention attached to transformer blocks.
BLIP-2~\cite{li2023blip2} tokenizes the image with a pretrained CLIP~\cite{zhong2022regionclip, EVA-CLIP}, which is later input into the language model with the bridge of an attention-based Q-former.
Instead of freezing the language model, Kosmos-1~\cite{huang2023language_kosmos} freezes the visual tokenizer while trains the language model from scratch with large-scale text and image-text data.  

Recently, with the open source of Large Language Models~\cite{touvron2023llama, vicuna, radford2019language_gpt2, chung2022scaling_flant5}, a lot of large multimodal models are constructed based on them. 
Mini-GPT4~\cite{zhu2023minigpt4} is built on the instruction-tuned Vicuna~\cite{vicuna} and the visual encoder from BLIP-2~\cite{li2023blip2}, with only a linear layer trained to bridge the two modules.
This simple design results in a powerful multi-modal chatbot, with noticeable vision-language understanding capability.
LLaVa~\cite{liu2023visual_llava} adopts CLIP as visual tokenizer, and trains the projector with a curated dataset with balanced concepts.
The model then can be finetuned for downstream tasks, \textit{e.g.}, ScienceQA~\cite{lu2022learn_scienceQA}. 
Apart from using frozen visual tokenizer, mPLUG-OWL~\cite{ye2023mplugowl} tunes the Perceiver Resampler with large-scale image-text data in the first stage, followed by the finetuning of language model with LoRA~\cite{hulora} in the second stage.
Although these generalist models have demonstrated impressive capability on multimodal tasks, we find that they mostly focus on the semantic understanding of the image, ignoring more fine-grained visual perception.
To tackle this incapability, we tune a new visual tokenizer with better fine-grained visual perception capabilities to further advance MLLMs as generalists.

\subsection{Visual Tokenizer Pretraining}
Visual encoders have been shown to benefit from large-scale pretraining for downstream tasks. 
The most common approach first pretrains the model on a large dataset with annotations, \textit{e.g.}, ImageNet~\cite{russakovsky2015imagenet}, and finetunes it for downstream tasks such as semantic segmentation~\cite{zhou2019semantic_ade20k} and object detection~\cite{lin2014microsoft_coco}. 
Recently, self-supervised pre-training have also shown to improve model's representation capability. 
The typical contrastive-based methods~\cite{caron2021emerging_dino, chen2020simple, chen2021exploring} trains the model by aligning views from the same image.
Inspired by the idea of mask-language-modeling for pretraining language models~\cite{kenton2019bert}, masked-image-modeling has also evolved for visual encoder pretraining.
These methods mask a proportion of image patches before feeding them into the model, and ask the model to recover the masked patches.
Some methods~\cite{baobeit} discretize the masked patches via a pretrained tokenizer~\cite{ramesh2021zero}, and ask the model to find the ID of the masked patch during pretraining.  
Besides, the momentum update of a model itself can also be used as an effective tokenizer~\cite{zhouimage_ibot, oquab2023dinov2}.
Recently, auto-encoder based~\cite{he2022masked_mae} methods ask the model to directly generate the masked patch in the continuous space.
Another stream of visual encoders is pretrained on massive image-text pairs via contrastive learning.
The most typical model CLIP~\cite{radford2021learning_clip} has been shown to be capable of various downstream tasks in zero-shot manner.
It has also evolved with more training data~\cite{openclip} and better optimization strategy~\cite{EVA-CLIP}.


\section{Discussions}

\subsection{Potential Societal Impacts}

\noindent\textbf{Potential Positive Impacts.}
In this work, we systematically investigated various visual pretraining methods under MLLM's framework.
Our findings may further motivate researchers in the community to design new visual pretraining algorithms.

\noindent\textbf{Potential Negative Impacts.}
The training process of large models often requires huge computation resources, which consumes a lot of energy and can exacerbate the emission of carbon dioxide. 
Furthermore, the training datasets may contain harmful contents, leading to biased prediction or harmful generation.

\subsection{Limitations}
In this work, our investigations are mostly based on released checkpoints, aiming to provide a guideline for researchers to select visual tokenizer accordingly.
Given that these models can be pretrained with different dataset and protocol, a more in-depth study could be performed by fully aligning their training procedure.

\section{Conclusion and Future Work}
We comprehensively studied various visual tokenizers through the lens of MLLM. 
Our investigation reveals that i) 
fully/weakly supervised models perform better than self-supervised ones on semantic representation, but this gap can be narrowed by scaling up pretraining dataset.
ii) Self-supervised models are better at fine-grained visual perception, where patch-level supervision is particularly effective.
iii) jointly tuning the visual tokenizer on the small-scale instruction dataset leads to the loss of rich semantics from large-scale pretraining.
Based on these findings, we hope to find a visual tokenizer that excels at both semantic understanding and fine-grained visual perception.
We reviewed existing methods and find that directly fine-tuning CLIP with region-supervision does not lead to a versatile visual tokenizer.
Besides, the mask strategy for pretraining is not suitable due to the train-test mismatch. 
Based on the insights above, we tune a new visual tokenizer, which distills CLIP patch feature into a new model without masking.
Equipped with our visual tokenizer, Vicuna can better understand images at multiple levels, results in superior performance on vision-language tasks including VQA, Image Captioning, Object Counting and Multi-Class Identification.
For future work, we would like to explore more versatile visual tokenizer that is capable of more challenging visual understanding tasks such as open-vocabulary object detection.

{\small
\bibliographystyle{unsrt}
\bibliography{egbib}
}

\newpage


\section*{A. Implementation Detail}

\subsection*{A.1 Implementation Detail for Empirical Studies}
For the experiments in empirical studies, we use a combination of 1) image captioning datasets: MS-COCO~\cite{lin2014microsoft_coco}, SBU~\cite{vicente2016large_sbu}, CC-3M~\cite{sharma2018conceptual_gcc} and Visual Genome~\cite{krishna2017vg} and 2) two object detection datasets, including Object365~\cite{shao2019objects365} and OpenImagesV6~\cite{kuznetsova2020openimages}. 
For image captioning data, we take the question \textit{"what does the image describe?"} as input prompt and ask the model to generate the descriptions.
For object detection datasets, we use a total of 6 tasks to fully utilize the rich annotations. Please refer to Section D in this appendix for more details.
The training dataset is uniformly sampled during training. 
We optimize the model with a learning rate of 1e-4 and a batch size 1024.
The whole model is optimized by the AdamW~\cite{loshchilov2017decoupled_adamw} optimizer and we set $\beta_1$ to 0.9 and $\beta_2$ to 0.98.
We train the model for 10k steps, while the learning rate is linearly warmuped from 0 in the first 1k steps, and is cosine decayed to 0 afterwards.
We optimize all models using \texttt{float16}.

\subsection*{A.2 Implementation Detail for GVT}
The implementation detail of our GVT is similar to that in the empirical studies, except that we use more data and more training steps.
Besides the image captioning and object detection dataset, we also used LLaVa-150k dataset~\cite{liu2023visual_llava}, which is generated by external powerful LLM.
We trained the model for 50k steps, with 2k steps for linear warmup. Then, we use cosine decay to decrease the learning rate to 0.

\subsection*{A.3 Evaluation Details.}

\noindent\textbf{VQA.} Modern Language Models mainly generate one or multiple sentences, making it infeasible to directly evaluate the MLLMs in the standard evaluation protocol which requires the prediction and ground truth to be exactly matched.
As such, we slightly relax the original evaluation protocol. We use the first sentence generated my MLLM as prediction result, and treated it as correct if \textit{contains} the ground truth answer.

\noindent\textbf{Image Captioning.}
When MLLMs generate multiple sentences, we use the first sentence as the captioning result for evaluation. 
Since MLLMs tend to generate multiple sentences, we use the prompt \textit{"Describe this image in a sentence: This is an image of"} as prompt to condense the prediction for effective evaluation.

\noindent\textbf{Object Counting.}
We extract the number of word from the first generated sentence, and compare it with ground truth number.

\noindent\textbf{Object Existence.}
We extract "yes" or "no" from the first generated sentence, and compare it with ground truth.

\begin{table}[ht]
    \centering
    \caption{Dataset set statistics of our dataset for evaluation.}\label{tab:label_stat}
    \begin{tabular}{c|c|c|c}
         \toprule
         Task & Split & Dataset & \# of Instance  \\
         \midrule
         Visual Question Answering  & validation & VQAv2~\cite{goyal2017making_vqav2} & 440k \\
         Image Captioning & validation & MS-COCO~\cite{lin2014microsoft_coco} & 25k \\
         Object Counting  & validation & MS-COCO~\cite{lin2014microsoft_coco} & 10k \\
         Object Counting  & validation & VCR~\cite{zellers2019recognition_vcr} & 10k \\
         Multi-Class Identification & validation & MS-COCO~\cite{lin2014microsoft_coco} & 10k \\
         Multi-Class Identification & valitdaion & VCR~\cite{zellers2019recognition_vcr} & 10k \\
         \bottomrule
    \end{tabular}
\end{table}

\section*{B. Benchmarking Fine-Grained Visual Understanding Tasks}\label{sec:region_benchmark}
We provide the details of the dataset used for evaluation in each task in Table~\ref{tab:label_stat}.
In this work, we constructed two fine-grained perception tasks: \textit{object counting} and \textit{object existence} based on instance-level annotations from existing datasets.
Specifically, they are constructed on MS-COCO~\cite{lin2014microsoft_coco} and VCR~\cite{zellers2019recognition_vcr} validation datasets.
We provide their details as follows.

\subsection*{B.1 Object Counting}
Besides the visual features, the prompt of this task -- \textit{"Question: How many \{obj\} are there in the image? Answer:\"} is fed into the MLLM for evaluation.
We select the object name \textit{\{obj\}} from the object list of the dataset.
Since there are often a single object of a certain class in one image, we select a maximum of 3 objects with highest occurrence in the image to make this benchmark challenging.
Similar to object counting benchmarks, we report Mean Absolute Error (MAE) and Root Mean Square Error (RMSE). 
Furthermore, we also report accuracy which treats the counting as a classification problem during evaluation.
Both COCO-OC and VCR-OC contain a total of 10k tasks.

\subsection*{B.2 Multi-Class Identification}
Multi-label classification can be used as task to evaluate the model's multi-instance understanding capability. 
However, given the open-ended nature of language models, the evaluation process is not stable since the language model may generate more fine-grained object names than the dataset categories, making a stable and fair evaluation difficult.
To this end, we change the format of this task and make the evaluation process more stable.
We design the prompt as \textit{"Question: Does \{obj\} exist in the image?" Answer:\"}, and the model is expected to answer \textit{"Yes"} or \textit{"No"}.
We select the object name \textit{\{obj\}} from the object list of the dataset.
For each image, we randomly select at most 3 objects that exist in the image, and the same number of objects that does not appear in the image, so as to make the evaluation set balanced. 
Both COCO-MCI and VCR-MCI contain a total of 10k tasks.

\section*{C. More Fine-grained Visual Understanding Results}
In this section, we provide more detailed results on our two new tasks: OC and OCI.

\noindent\textbf{Detailed Object Counting Results.}
We show the detailed results of Object Counting task on MS-COCO in Table~\ref{tab:count_coco}.
It can be observed that, when the images contains relatively small number of objects (1-3), all methods can understand the number of objects to some extend, where ours is significantly better than others.
However, when the images become more complex, where the number of occurrence increases (4-6, 7-9), the performance has significantly dropped. 
Similar trend can also observed in Table~\ref{tab:count_vcr}.  
These results demonstrate that current MLLMs still struggle at correctly counting the objects, indicating future research are required to make them more capable of challenging visual understanding tasks.

\begin{table}[ht]
    \caption{Detailed results on the Object Counting on MS-COCO dataset.}\label{tab:count_coco}
    \scalebox{0.70}{
    \begin{tabular}{l|ccc|ccc|ccc|ccc}
    \toprule
    GT range & \multicolumn{3}{c|}{1 - 3} & \multicolumn{3}{c|}{4 - 6} & \multicolumn{3}{c|}{7 - 9} & \multicolumn{3}{c}{Overall} \\ 
    Method & Acc $\uparrow$ & MAE $\downarrow$ & RMSE $\downarrow$ & Acc $\uparrow$ & MAE $\downarrow$ & RMSE $\downarrow$  & Acc $\uparrow$ & MAE $\downarrow$ & RMSE $\downarrow$ & Acc $\uparrow$ & MAE $\downarrow$ & RMSE $\downarrow$ \\
    \hline
    MiniGPT4 & 23.0 & 0.96 & 1.60 & 11.0 & 1.68 & 2.19 & 0.0 & 4.09 & 4.24 &21.1 & 1.36 & 2.1 \\
    LLaVa & 26.5 & 0.89 & 1.86 & 11.0 & 1.72 & 3.25 &1.58 & 4.75 & 5.83 & 22.0 & 1.36 & 2.70 \\
    BLIP-2 & 61.1 & 0.47 & 0.82 & 12.1 & 2.10 & 2.50 & 0.47 & 4.97 & 2.57 & 48.0 & 1.15 & 2.05 \\
    \midrule
    GVT (Ours) & 74.7 & 0.25 & 0.51 & 4.7 & 2.26 & 2.49 & 0.02 & 2.29 & 5.25 & 56.0 & 1.01 & 1.93 \\
    \bottomrule
    \end{tabular}
    }
\end{table}

\begin{table}[ht]
    \caption{Detailed results on the Object Counting on VCR dataset.}\label{tab:count_vcr}
    \scalebox{0.70}{
    \begin{tabular}{l|ccc|ccc|ccc|ccc}
    \toprule
    GT range & \multicolumn{3}{c|}{1 - 3} & \multicolumn{3}{c|}{4 - 6} & \multicolumn{3}{c|}{7 - 9} & \multicolumn{3}{c}{Overall} \\ 
    Method & Acc $\uparrow$ & MAE $\downarrow$ & RMSE $\downarrow$ & Acc $\uparrow$ & MAE $\downarrow$ & RMSE $\downarrow$  & Acc $\uparrow$ & MAE $\downarrow$ & RMSE $\downarrow$ & Acc $\uparrow$ & MAE $\downarrow$ & RMSE $\downarrow$ \\
    \hline
    MiniGPT4 & 25.0 & 0.84 & 1.32 & 13.0 & 1.48 & 1.82 & 0.00 & 4.34 & 4.46 & 25.0 & 1.51 &2.24 \\
    LLaVa & 24.0 & 0.91 & 2.24 & 13.3 & 1.53 & 1.99 & 1.16 &4.46 & 4.75 & 24.0 & 1.58 & 2.77 \\
    \midrule
    GVT (Ours) & 63.9 & 0.36 &0.61 & 5.94 & 2.22 & 2.46 & 0.00 & 4.96 & 5.18 & 40.0 & 1.49 & 2.41 \\
    \bottomrule
    \end{tabular}
    }
\end{table}

\noindent\textbf{Detailed Multi-Class Identification Results.}
We provide more detailed results on MCI task for MS-COCO in Table~\ref{tab:mci_coco}.
The performance of all methods decrease when the image becomes more complex (with more objects in the image). 
However, the results on the VCR dataset does not show a stable trend.
We conjecture this can be related to the difference on the instruction tuning datasets, which leads the model to focus on different types of objects.

\begin{table}[ht]
    \centering
    \caption{Detailed results on the Multi-Class Identification on MS-COCO dataset.}\label{tab:mci_coco}
    \begin{tabular}{l|ccc|c}
    \toprule
    \#Objects & {1 - 9} & 10 - 20 &  > 20 & Overall \\ 
    \midrule
    MiniGPT4 & 80.7 & 72.3 & 96.1 & 76.8 \\
    LLaVa & 52.1 & 52.0 & 51.7 & 52.0 \\
    BLIP-2 & 85.4 & 77.6 & 75.2 & 81.9 \\
    \midrule
    GVT (Ours) & 89.7 & 87.0 & 84.5 & 88.2 \\
    \bottomrule
    \end{tabular}
\end{table}

\begin{table}[ht]
    \centering
    \caption{Detailed results on the Multi-Class Identification on VCR dataset.}\label{tab:mci_vcr}
    \begin{tabular}{l|ccc|c}
    \toprule
    GT range & {1 - 9} & 10 - 20 &  > 20 & Overall \\ 
    \midrule
    MiniGPT4 & 71.2 & 70.2 & 71.1 & 70.8 \\
    LLaVa & 67.1 & 66.6 & 66.8 & 66.9 \\
    BLIP-2 & 67.6 & 70.3 & 70. & 68.9 \\
    \midrule
    GVT (Ours) & 77.1 & 80.6 & 81.5 & 78.8 \\
    \bottomrule
    \end{tabular}
\end{table}




\section*{D. Object-centric Tasks}\label{sec:obj_task}
The work of~\cite{piergiovanni2022pre_objaware} has proposed 4 tasks to utilize object detection dataset for vision-language pretraining, including:

\noindent\textbf{1. List Objects} \\
    Input: \textit{"List all objects"} \\
    Output: \textit{"\{obj1\}, \{obj2\}, ..."} \\

\noindent\textbf{2. Object Existence} \\
    Input: \textit{"Does \{obj\} exist in the image?"} \\
    Output: \textit{"Yes/No."} \\

\noindent\textbf{3. Group Existence} \\
    Input: \textit{"Does all of \{obj1\}, \{obj2\} and \{obj3\} exists in the image?"} \\
    Output: \textit{"Yes/No."} \\

\noindent\textbf{4. Existence Selection} \\
    Input: \textit{"Which of \{obj1\}, \{obj2\}, \{obj3\} exist in the image?"} \\ 
    Output: \textit{"\{obj1/2/3\}"} \\

To further utilize the rich annotations in object detection datasets, we also design two tasks which facilitate the model's learning on fine-grained visual information. \\

\noindent\textbf{5. Object Counting} \\
    Input: \textit{"How many \{obj\}s are there in the image?"} \\
    Output: \textit{1-9}. \\

\noindent\textbf{6. Spatial Relation} \\
    Input: \textit{“What is the spatial relation between \{obj1\} and \{obj2\}? Choose one from Top/Top Left/Left/Bottom Left/Bottom/Bottom Right/Right/Top Right"} \\
    Output: "\textit{Top/Top Left/Left/Bottom Left/Bottom/Bottom Right/Right/Top Right}"

Task 6 is only performed when the selected \textit{\{obj1\}} and \textit{\{obj2\}} are unique in the image, so as to avoid the referring ambiguity problem.
For all tasks, we use the input text as the prompt and ask the model to generate the output text. The loss is only computed on the output texts.
For each image, the task is uniformly sampled on the two object detection datasets~\cite{shao2019objects365, kuznetsova2020openimages}.

\end{document}